\documentclass[submission,copyright,creativecommons]{eptcs}
\providecommand{\event}{SLPCS 2016} 
\usepackage{tikz}
\usepackage{pgfplots}
\usepackage{caption}
\usepackage{subcaption}
\usepackage{xfrac}
\usepackage{pifont}

\usetikzlibrary{shapes.arrows}

\newcommand{\cmark}{\ding{51}}%
\newcommand{\xmark}{\ding{55}}%

\title{Words, Concepts, and the Geometry of Analogy}
\author{Stephen McGregor \qquad\qquad Matthew Purver \qquad\qquad Geraint Wiggins
\institute{Queen Mary University of London}
\institute{School of Electronic Engineering and Computer Science}
\email{s.e.mcgregor@qmul.ac.uk \qquad m.purver@qmul.ac.uk \qquad geraint.wiggins@qmul.ac.uk}
}
\def\titlerunning{Words, Concepts, and the Geometry of Analogy}
\def\authorrunning{S. McGregor, M. Purver \& G. Wiggins}
\begin{document}
\maketitle

\begin{abstract}
This paper presents a geometric approach to the problem of modelling the relationship between words and concepts, focusing in particular on analogical phenomena in language and cognition.  Grounded in recent theories regarding geometric conceptual spaces, we begin with an analysis of existing static distributional semantic models and move on to an exploration of a dynamic approach to using high dimensional spaces of word meaning to project subspaces where analogies can potentially be solved in an online, contextualised way.  The crucial element of this analysis is the positioning of statistics in a geometric environment replete with opportunities for interpretation.
\end{abstract}

\section{Concepts in the Kingdom of Context}
``Laugh to scorn the power of man, for none of woman born shall harm Macbeth''---so says a phantasmal apparition to the ill-fated anti-hero of Shakespeare's gruesome tragedy.  And, taking the ghost at its word, Macbeth goes merrily on his bloodthirsty, tyrannical way until one day a man born by caesarean section chops him down.  What's at play here is, of course, the context through which the utterance has been interpreted: Macbeth made the classic mistake of taking a demonic entity literally rather than spending a little time thinking about malicious ways in which words describing a seemingly impossible scenario might actually be saying something imminently plausible.

In as much as theoretical linguistics is concerned, context, much more than just the stuff of literary misapprehensions, has become more or less the central problem in terms of understanding the way that words relate to situations in the world.  Whether it be type theory \cite{Montague1974,Cooper2012}, event semantics \cite{Davidson1969}, relevance theory \cite{SperberEA1995}, or any of a variety of other theoretical approaches to semantics, the ability to map word meaning to a specific mental experience hinges on the situatedness of both cognition and language.  What is at stake here is the fraught relationship between words and concepts.  Barsalou asks, "What kinds of representations are sufficiently expressive to account for the hierarchical conceptual relationships of concepts?", \cite{Barsalou1993}, and concludes that, just as the labelling of conceptual content is \emph{haphazard}, so language must be characteristised by a fundamental \emph{vagary} and \emph{flexibility}.

Yet there is something in the contextualisation of meaning that confounds computational approaches to language.  Symbol manipulating machines, bound by the literalness of the formal languages in which they traffic, balk at handling the immensity of context that can be associated with any given symbol in the course of natural communication.  An attempt to build robust symbolic representations linking linguistic elements to all their potential contextualised applications more or less immediately encounters a \emph{frame problem} by which a preponderance of potential contexts explodes with even very slight shifts in the situation of a cognitive-linguistic agent in the real world \cite{Dennett1984}.  The question, then, is how computers might use low-level statistical type representations to capture the environmentally situated conceptual dynamism that is essential to cognition.

G\"{a}rdenfors has proposed a theory of \emph{conceptual spaces} which moves conceptual modelling into a spatial domain that should at least in principle be more tractable to symbol manipulating machines \cite{Gardenfors2000}.  By describing concepts as being situated in spaces where their dimensions can be interpreted as properties of the concept, this theory offers hope that concepts might be modelled in terms of the kind of matrix of quantifications that are native to computers.  Widdows similarly proposes that concepts can be understood as geometric phenomena, focusing more explicitly on the relationship between concepts and words and using statistical approaches to lexical modelling to build a framework for logical operations on conceptual representations \cite{Widdows2004}.  In both cases, these approaches to concept modelling offer a kind of interstitial layer of information processing that sits between the low level data of environmental stimuli or textual statistics and the high level symbolism of cognitive perception.

This existing work provides a theoretical grounding for the idea that will be explored in the following pages, where the groundwork for developing a methodology for moving from word counts to conceptual models will be laid.  This paper will address a particular type of conceptual relationship: analogy, a conceptual mechanism with a pedigree dating back at least to Aristotle \cite{Aristotle1895}.  The utility of analogy as a mechanism for performing conceptualisation in scientific domains has been thoroughly attested \cite{Hesse1963}, and more generally analogy is inherent in the relationships that can be mapped across the hierarchical structure of a taxonomy.  So a move from the statistically proximate to the geometrically analogical is a move from the lexical to the conceptual.  The crucial aspect of the approach which will be proposed here is its geometry: by maintaining spaces that, in the spirit of G\"{a}rdenfors and Widdows, are strictly interpretable right down to the level of a single dimension, we hope to be able to eventually establish a robust computational process for dynamically extracting context sensitive conceptual mappings from large scale textual corpora.

Our dynamic model for projecting context specific, conceptually productive subspaces should likewise be situated in terms of other ongoing technical work in distributional semantics and further afield.  In particular, a similar approach has been explored for the computational discovery of taxonomic relationships such as hypernymy and hyponymy \cite{Rimell2014}, and likewise for the generation of spaces where directions in space can be understood semantically \cite{DerracEA2014}.  A key feature of each of these approaches is the establishment of a distributional semantic space where the preservation of statistical information about an underlying corpus yields an element of interpretability to the space.  More generally, work on knowledge base construction and extension has likewise treated geometric properties of spaces of entities as translations which can be passed between pairs of entities and likewise be interpreted as meaningful features of the space \cite{BordesEA2013}.

These approaches involving the translation of statistical information about a corpus into spaces where geometry coincides with conceptual properties should be contrasted with theories that frame concepts in terms of rule based manipulations of content-bearing symbols \cite{FodorEA1988}.  The case to be made here is, broadly, that cognition is an inherently representational process, anchored in the intentionality of the symbols in which it traffics rather than in the geometry of conceptual spaces.  Ultimately, though, there is hope that the geometrical approach can actually stand in as a construct mitigating between low level stimuli and high level symbolic representation.  There is likewise a point of comparison with work where conceptual relationships are modelled as sets of nodes in a semantic network \cite{Haspelmath2003}.  The latter type of model is characterised by abstract graph theoretical representations where relationships, signified by labelled edges, are inherently symbolic and geometry is not a factor in the interpretation of the semantic situation.  Having said this, it must also be noted that there is considerable scope for reconciling graph based approaches to conceptual modelling to the geometric methodology that will be endorsed here \cite{Gardenfors2014}.

\section{Spaces of Meaning}
A prevalent paradigm in corpus-based computational approaches to lexical modelling is \emph{distributional semantics}, which seeks to establish spaces where words are treated as vectors and the relative proximity of vectors corresponds to the semantic relatedness of the words that label those vectors.  In particular, the \emph{distributional hypothesis} holds that ``words that occur in similar contexts tend to have similar meanings,'' \cite[p. 148]{TurneyEA2010}, and so the intuition behind distributional semantics is that if the typical textual context of word types can somehow be statistically captured, then there is hope for constructing computationally tractable representations of word meaning based on word-vectors populated by these statistics.  Traditional approaches have utilised the frequencies or rates of word co-occurrences to build distributional semantic spaces \cite{SaltonEA1975,Schutze1992}, while more recent work has often employed non-linear regression to learn an optimally arranged space \cite{BengioEA2003,CollobertEA2008} or used matrix factorisation to optimise the informativeness of condensed spaces \cite{BaroniEA2010},though, as will be seen in the following analysis, the geometrical character of these spaces can vary.  Clark provides a current and comprehensive overview of the field \cite{Clark2015}.

Of particular relevance to the ideas presented in the present paper is the recent development of the \texttt{word2vec} distributional semantic model, which uses a neural network to learn a space of word meanings based on iterative traversals of a large scale corpus \cite{MikolovEA2013}.  Rather than building up statistical representations of word co-occurrences where the number of features in a space can be on the scale of or even significantly larger than the set of word-vectors constituting the space's vocabulary of word-vectors, this model gradually learns a space which ultimately resembles the statistically derived spaces.  The network approach to distributional semantics uses a set of arbitrary dimensions to gradually pull the space into order, further utilising negative sampling to push word-vectors that do not bear semantic relatedness apart.

The upshot of this approach is a space which yields some interesting semantic properties which at first appear to move towards the modelling of conceptual relationships.  In particular, it has been demonstrated that in \texttt{word2vec} spaces, straightforward linear algebraic procedures can be used to map analogical relationships between words.  The paradigmatic example offered in the literature is this:

\begin{equation}
\overrightarrow{woman}-\overrightarrow{man}+\overrightarrow{king} \approx \overrightarrow{queen}
\end{equation}

To put it slightly differently, the vector which projects from the point in space representing \emph{man} to the point in space representing \emph{woman} is roughly equal in both length and direction to the vector going from \emph{king} to \emph{queen}---or, to attempt to interpret this geometrically, these four word-vectors very closely form a parallelogram sitting in the high-dimensional space of \texttt{word2vec}.  The robustness of this methodology for analogy completion has been demonstrated on a fairly large dataset consisting of some tens of thousands of analogies that are input as \sfrac{3}{4} complete, with the model returning the correct fourth component of the analogy more than half the time \cite{MikolovEA2013b}.  It is worth noting that similar results have subsequently been reported using a model built through a combination of neural networks and matrix factorisation \cite{PenningtonEA2014}, and it has been suggested that the strong performance of \texttt{word2vec} might be down to the fine tuning of a set of hyperparameters which could likewise be implemented in a more traditional statistical distributional semantic model \cite{LevyEA2014}.

Nonetheless, it is notable, if not outright surprising, that these types of ostensibly geometric relationships should emerge from a highly non-linear analysis of co-occurrence relationships across a large scale corpus.  In particular, at this point, the relationships at least seem to veer away from the purely lexical and into the conceptual.  By capturing the relationship between word-vectors, the model begins to flesh out the corresponding relationships between correspondingly labelled concepts: so, for instance, we discover that in the conceptual domain of \textsc{royalty}, there is a dimension of \textsc{gender}.  By systematically going through the ostensible geometry of the space, we might imagine gradually building up a network of relationships resembling a kind of taxonomy, where the coherence between labels across domains offers the trace of a space of interrelated ideas about things in the world.

A closer inspection of the model's operation, however, reveals some significant caveats.  For one thing, the dataset used in the evaluation of the model's analogy completing performance is large, but tends to feature analogies with a distinct character: of the analogies described as ``semantic'', which is roughly half of them, most involve proper names, and a subset involve familial relationships.  Then about half the test set is designed to explore ``syntactic'' relationships, and these analogies mainly consist of of comparisons between morphologically aligned pairs of words.  In the case of proper names, the analogical queries are arguably less hindered by the ambiguity that characterises more general usage.  And in the case of the more grammatically oriented analogies, the model might be capturing the fact that word stems are associated with a general semantic context, but specific derivational words are also characterised by a particular grammatical context---with, for example, adjectives more likely to occur in the immediate context of nouns, where adverbs are to be found in the context of verbs.

In order to draw out this point a bit, here are six examples of analogies completed by \texttt{word2vec}.  In each case, the model is offered three input terms, and it finds a fourth term (underlined) based on the linear algebraic operation between vectors described above.  For the sake of analysis, three examples have been picked which do seem intuitively to work, and three which don't seem to work:

\begin{enumerate}
\item \emph{Paris} is to \emph{France} as \emph{Finland} is to \underline{\emph{Helsinki}} \cmark
\item \emph{girl} is to \emph{boy} as \emph{prince} is to \underline{\emph{princess}} \cmark
\item \emph{fast} is to \emph{faster} as \emph{big} is to \underline{\emph{bigger}} \cmark
\item \emph{elephant} is to \emph{mouse} as \emph{big} is to \underline{\emph{huge}} \xmark
\item \emph{bank} is to \emph{river} as \emph{shoulder} is to \underline{\emph{elbow}} \xmark
\item \emph{picture} is to \emph{paint} as \emph{story} is to \underline{\emph{mexican\_viagra\_viagra}} \xmark
\end{enumerate}

In (4), the model seems to stumble on the conceptually crucial notion of antonymy: in a lexical model based on word co-occurrences, words with opposite meanings are actually very likely to occur in similar contexts, so there is not a directed way to know about the difference between scaling symmetrically between \textsc{largeness} and \textsc{smallness} versus scaling up within the domain of \textsc{largeness}.  Likewise in (5), the model evidently stumbles on a classic example of lexical ambiguity, with both \emph{bank} and \emph{shoulder} being used in a contextually specific way that's perfectly clear to an attuned speaker of the language but that proves inexorably muddled for a model that is only aware of morphological statistics.  And in the case of (6), where a plausible completion of the relationship would have been something along the lines of \emph{words} (or perhaps, more idiomatically, \emph{tell}), the model has completely failed to capture the metaphoric aspect of this particular analogy, to moderately humorous effect.

In each of the cases offered, the model is seen to fall short when it comes to the essential contextuality of the analogy being mapped.  This belies the fundamentally non-geometric characteristic of this type of space: distances corresponding to word similarity can, on their own, occasionally be useful heuristics for predicting more nuanced conceptual relationships denoted by words, but in general this space of word statistics can't really be mapped to a more productive conceptual space where dimensions take on meaningful properties of their own.  In order to move from spaces of words to spaces of concepts, we need to discover a more dynamic way of manipulating the space itself, mirroring the fundamental situatedness of the way that a cognitive agent conceptualises about the world.

\section{Putting Word Context in Context}
So it is evident, and perhaps, from a theoretical perspective, already somewhat obvious, that context is essential in the mapping of the relationship between words and concepts.  The problem with the lexical spaces described thus far, however, is that, while they are very effective at capturing a kind of encyclopedic knowledge about the relationships between words and concepts, they are also essentially \emph{static}.  In the case of \texttt{word2vec} in particular, the neural network method used to generate the space is specifically designed to maximise the utility of each of the spaces dimensions, treating the dimensions themselves as nothing more than handles to gradually shift the space into order based on iterative observations of words in context across a corpus.  And in the case of other state-of-the-art systems, the use of matrix factorisation techniques such as SVD results in the optimisation of the informativeness of each dimension in the space, but at the expense of the degradation of the informativeness of the dimensions themselves.  Thus these spaces are \emph{static} in the sense that there is no real hope of finding a mechanism for further interpreting and manipulating their dimensions in a context sensitive way.

\begin{figure}[t]
\centering
\begin{subfigure} [t] {0.45 \textwidth}
	\begin{tikzpicture}[scale=0.07,baseline]
		\draw (0,0)--(-100,0)--(-100,-100)--(0,-100)--(0,0);

    	\draw (-52,-52) [above] node {cat};
        \fill (-52,-52) circle[radius=1];

    	\draw (-55,-20) [above] node {lion};
        \fill (-55,-20) circle[radius=1];

    	\draw (-25,-48) [above] node {house};
        \fill (-25,-48) circle[radius=1];

    	\draw (-48,-25) [above] node {tiger};
        \fill (-48,-25) circle[radius=1];
        
        \draw (-45,-75) [above] node {jungle};
        \fill (-45,-75) circle[radius=1];
        
        \draw (-80,-55) [above] node {hamster};
        \fill (-80,-55) circle[radius=1];
        
        \draw (-95,-45) [above] node {dog};
        \fill (-95,-45) circle[radius=1];

        \node at (-50,-102.5) [single arrow,draw,rotate=90,minimum height=40,minimum width=40,inner sep=15,opacity=0] {};
    \end{tikzpicture}
\end{subfigure}
\hspace{0.03 \textwidth}
\begin{subfigure}[t]{0.45 \textwidth}
	\begin{tikzpicture}[scale=0.07,baseline]
    	\draw (0,0)--(-100,0)--(-100,-100)--(-57.5,-100);
        \draw (-42.5,-100)--(0,-100)--(0,-57.5);
        \draw (0,-42.5)--(0,0);

    	\draw (-52,-52) [above] node {cat};
        \fill (-52,-52) circle[radius=1];
        \draw[dashed,->] (-52,-52)--(-52,-100);
        \draw (-52,-100) node {x};
        \draw[dashed,->] (-52,-52)--(0,-52);
        \draw (0,-52) node [rotate=90] {x};

    	\draw (-55,-20) [above] node {lion};
        \fill (-55,-20) circle[radius=1];
        \draw[dashed,->] (-55,-20)--(-55,-100);
        \draw (-55,-100) node {x};
        \draw[dashed,->] (-55,-20)--(0,-20);
        \draw (0,-20) node [rotate=90] {x};

    	\draw (-25,-48) [above] node {house};
        \fill (-25,-48) circle[radius=1];
        \draw[dashed,->] (-25,-48)--(-25,-100);
        \draw (-25,-100) node {x};
        \draw[dashed,->] (-25,-48)--(0,-48);
        \draw (0,-48) node [rotate=90] {x};

    	\draw (-48,-25) [above] node {tiger};
        \fill (-48,-25) circle[radius=1];
        \draw[dashed,->] (-48,-25)--(-48,-100);
        \draw (-48,-100) node {x};
        \draw[dashed,->] (-48,-25)--(0,-25);
        \draw (0,-25) node [rotate=90] {x};
        
        \draw (-45,-75) [above] node {jungle};
        \fill (-45,-75) circle[radius=1];
        \draw [dashed,->] (-45,-75)--(-45,-100);
        \draw (-45,-100) node {x};
        \draw [dashed,->] (-45,-75)--(0,-75);
        \draw (0,-75) node [rotate=90] {x};
        
        \draw (-80,-55) [above] node {hamster};
        \fill (-80,-55) circle[radius=1];
        \draw [dashed,->] (-80,-55)--(-80,-100);
        \draw (-80,-100) node {x};
        \draw [dashed,->] (-80,-55)--(0,-55);
        \draw (0,-55) node [rotate=90] {x};
        
        \draw (-95,-45) [above] node {dog};
        \fill (-95,-45) circle[radius=1];
        \draw [dashed,->] (-95,-45)--(-95,-100);
        \draw (-95,-100) node {x};
        \draw [dashed,->] (-95,-45)--(0,-45);
        \draw (0,-45) node [rotate=90] {x};
        
        \node at (3,-50) {\Huge\}};
        \node at (-50,-103) [rotate=-90] {\fontsize{80pt}{4em}\selectfont \}};
        
        \node at (-50,-102.5) [single arrow,draw,rotate=90,minimum height=40,minimum width=40,inner sep=15] {};
        \node at (-50,-107) {\textsc{beast}};
        \node at (2.5,-50) [single arrow,draw,rotate=180,,minimum height=40,minimum width=40,inner sep=15] {};
        \node at (7,-50) [rotate=90] {\textsc{pet}};
    \end{tikzpicture}
    \label{fig:redux}
\end{subfigure}
\caption{Contextual Projections.  On the left, a base space exhibits vagary regarding the concept \textsc{cat}, captured by the proximity of the word-vector $\protect\overrightarrow{cat}$ to terms associated with different connotations of the word \emph{cat}.  On the right, two different perspectives on this space reveal context specific conceptual clusterings.}
\label{FIG:perspectives}
\end{figure}
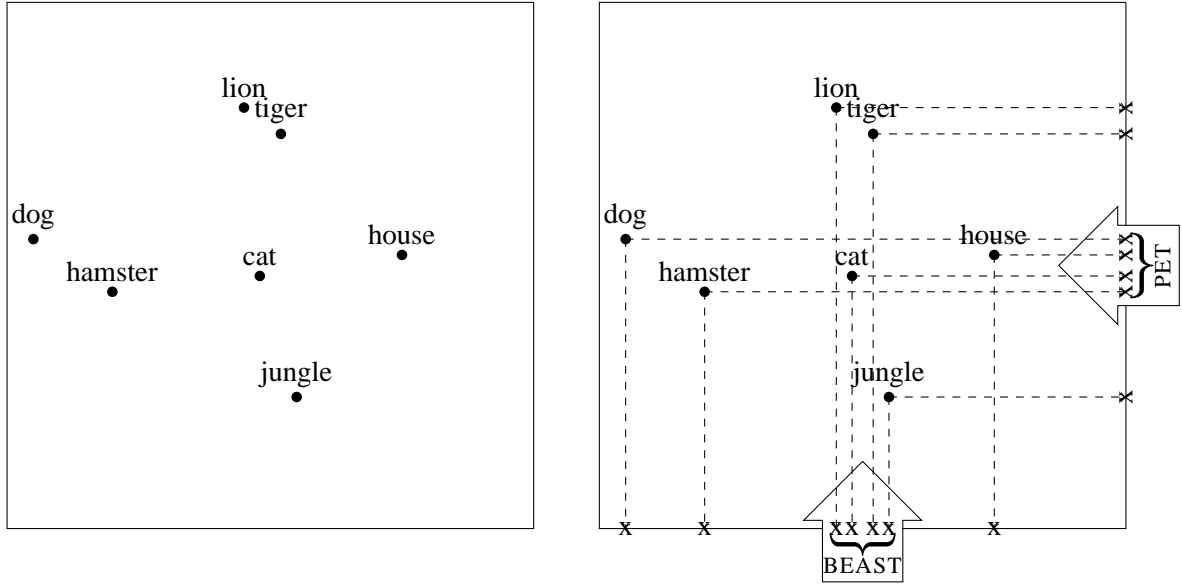

Recent work has suggested an alternative approach to distributional semantics, by which a very high dimensional space of word co-occurrence statistics is learned and left unrefined.  This base space can then take sets of words as input and analyse the corresponding word-vectors in order to determine the statistically literal co-occurrence dimensions which best capture the conceptual context of the relationship between the input words.  The sparsity of the resulting space becomes its strength, affording the projection of a subspace in which terms become clustered in a conceptually coherent way: the base space is \emph{dynamic} in that it responds to input in an online, situational fashion.  The intuition behind this approach is illustrated in Figure~\ref{FIG:perspectives}.  Furthermore, the approach has been implemented and its conceptual clusterings verified by human subjects \cite{AgresEA2015}, in addition to being shown to perform at least equivalently to \texttt{word2vec} on a computational task involving contextualising inputs in the form of compound nouns \cite{McGregorEA2015c}.  It should also be noted that a similar approach has been explored for the computational discovery of taxonomic relationship such as hypernymy and hyponymy \cite{Rimell2014}.

The novel proposal offered in the present paper is that a similarly context sensitive approach involving online projections from a base space should be applicable to the task of generating subspaces where analogical relationships can be mapped in a robustly geometric way.  It stands to reason to assume that such a dynamic model should, in principle, be able to find a way, given an analogy, to project some subspace where the analogy can be described as a parallel relationship between paired vectors.

But, whereas the published work mentioned above analyses input by searching for dimensions that maximise the values of a set of input terms, the way forward for projecting analogically productive subspaces must lie in searching for dimensions that optimise the desired geometric relationship between the components of the analogy.  To express this formulaically, given an analogy \emph{A} is to \emph{B} as \emph{C} is to \emph{D}, dimensions should be sought were the values of the analogical components play out such that $A-B \approx C-D$.  The hypothesis offered here, as a preliminary step towards an approach to analogy completion and perhaps even metaphor generation, is that dimensions selected on this basis should produce spaces where the previously described geometric relationship between analogical terms emerges.

In order to test this hypothesis and further explore the nature of analogically geared subspaces, the following procedure is proposed, based on an input consisting of an analogy of the form $A:B::C:D$.  First, in line with existing work in contextually dynamic distributional semantics \cite{AgresEA2015,McGregorEA2015c}, a general base space of co-occurrence statistics is built---in particular, the space used to generate the following data is based on the English language pages of Wikipedia, and the statistical metric employed is pointwise mutual information measuring the likelihood of words co-occurring within 5 words of one another.  Specifically, a matrix of word-vectors is calculated as follows, where $n_{w,c}$ is the frequency of observed co-occurrences of context word $c$ within a 5-word window of target word $w$, $n_w$ and $n_c$ are the independent frequencies of $w$ and $c$, $W$ is the total number of occurrences of words within the vocabulary, and $a$ is a smoothing constant (set in this case to 10,000, a value determined experimentally):

\begin{equation}
M_{w,c} = \log_2 \left(\frac{n_{w,c} \times W}{n_w \times \left(n_c + a\right)} + 1\right)
\end{equation}

The result is a sparse space consisting of a vocabulary of word vectors representing the 200,000 most frequent words in Wikipedia arrayed across dimensions representing co-occurrences with approximately 7.5 million word types.  With this base space established, the four components of the analogy are analysed, and an initial subspace $N$ of the $d$ dimensions that have non-zero values for the word vectors associated with A, B, and C is built.  This dense $3 \times d$ space is then normalised using L1 normalisation, ie, so that the values of each word-vector sum to 1.  Next, the 200 dimensions with the highest mean values are selected.  In other words, we select the dimensions with the highest value for the following equation:

\begin{equation}
\mu_c =  \frac{1}{3} \sum_{w \in \{A,B,C\}} M_{w,c}
\end{equation}

This step generates a subspace of dimensions that are particularly relevant to the context of the analogy being examined.  Furthermore, the normalisation of the $3 \times d$ matrix of non-zero values ensures that a word-vector characterised by dimensions with lower values will still be proportionally represented in the next step of the selection of an analogical context.  It's worth noting, as well, that the dimensions selected here are chosen only based on their relevance to the first three components of the analogy, so there is already an element of restriction here: there is no guarantee that component D will be represented with a non-zero value on any of these dimensions.  With an eye towards eventually developing a robust methodology for analogy completion, a strong outcome using this procedure guarantees that there is some hope of using only the information afforded by A, B, and C to find an analogical subspace where A:B::C:D is mapped as a predictable geometric relationship.

Next, from these 200 dimensions, the 20 dimensions $c$ with the lowest values for the equation $((M_{A,c}-M_{B,c})-(M_{C,c}-M_{D,c}))^2$ are selected.  This 20-dimensional subspace is then considered the context in which the analogy is mapped.  

To demonstrate the robustness of this approach, which is to say, to show that there is almost always \emph{some} subspace where a given analogy can be mapped geometrically, a portion of the \texttt{word2vec} analogy testset consisting of 1,000 analogies has been held back, and a 20 dimensional subspace has been determined for each analogy following the procedure described above.  In each case, the space has then been examined to see if the previously attested linear algebraic method for analogy completion was employed, leaving out the fourth term in each case and instead returning the word-vector nearest to the point described by the equation $\overrightarrow{B}-\overrightarrow{A}+\overrightarrow{C}$.  Of the 1,000 anlogies tested, 3 contained words not in our model's 200,000 word vocabulary; of the remaining 997, only one was not completed correctly.

\begin{figure}
\centering
\begin{subfigure} [t] {0.3 \textwidth}
\begin{tikzpicture}[scale=0.5]
	\begin{axis} [xlabel=rank,ylabel=PMI]
	\addplot [mark=none] coordinates {
		(0,0.02327)(104,0.09666)(208,0.12587)(312,0.15134)(416,0.16899)(520,0.18883)(624,0.20704)(728,0.22586)(832,0.2433)(936,0.25857)(1040,0.27579)(1144,0.2935)(1248,0.31131)(1352,0.32865)(1456,0.3435)(1560,0.35825)(1664,0.37511)(1768,0.39098)(1872,0.40593)(1976,0.42345)(2080,0.43902)(2184,0.4532)(2288,0.46775)(2392,0.48252)(2496,0.49892)(2600,0.51413)(2704,0.5304)(2808,0.54273)(2912,0.55435)(3016,0.56793)(3120,0.58291)(3224,0.59749)(3328,0.61382)(3432,0.62895)(3536,0.64204)(3640,0.65608)(3744,0.67218)(3848,0.68741)(3952,0.70354)(4056,0.71704)(4160,0.73224)(4264,0.74858)(4368,0.76273)(4472,0.77758)(4576,0.79366)(4680,0.80654)(4784,0.82164)(4888,0.83789)(4992,0.85428)(5096,0.86902)(5200,0.88407)(5304,0.89873)(5408,0.91574)(5512,0.93367)(5616,0.94932)(5720,0.96576)(5824,0.98529)(5928,0.99945)(6032,1.01544)(6136,1.03199)(6240,1.04698)(6344,1.06385)(6448,1.0779)(6552,1.09863)(6656,1.11679)(6760,1.13618)(6864,1.1568)(6968,1.17395)(7072,1.19348)(7176,1.21061)(7280,1.22893)(7384,1.24539)(7488,1.26312)(7592,1.28549)(7696,1.30581)(7800,1.32566)(7904,1.34743)(8008,1.36918)(8112,1.39024)(8216,1.41266)(8320,1.43216)(8424,1.45023)(8528,1.46731)(8632,1.49015)(8736,1.51154)(8840,1.53226)(8944,1.55345)(9048,1.57278)(9152,1.59295)(9256,1.61294)(9360,1.63483)(9464,1.657)(9568,1.68201)(9672,1.70474)(9776,1.72666)(9880,1.74766)(9984,1.76973)(10088,1.79548)(10192,1.81861)(10296,1.83896)(10400,1.86226)(10504,1.88308)(10608,1.90952)(10712,1.93541)(10816,1.95746)(10920,1.98038)(11024,2.0086)(11128,2.03564)(11232,2.05789)(11336,2.08379)(11440,2.10837)(11544,2.13383)(11648,2.15788)(11752,2.18577)(11856,2.21097)(11960,2.23709)(12064,2.26572)(12168,2.29662)(12272,2.32831)(12376,2.35274)(12480,2.37908)(12584,2.40856)(12688,2.43758)(12792,2.46818)(12896,2.50012)(13000,2.52761)(13104,2.55764)(13208,2.5846)(13312,2.61558)(13416,2.64493)(13520,2.67546)(13624,2.71447)(13728,2.74444)(13832,2.77802)(13936,2.81666)(14040,2.85477)(14144,2.88693)(14248,2.92487)(14352,2.95442)(14456,2.99591)(14560,3.03481)(14664,3.07175)(14768,3.1103)(14872,3.14432)(14976,3.17544)(15080,3.21722)(15184,3.25313)(15288,3.29221)(15392,3.32931)(15496,3.36406)(15600,3.40515)(15704,3.44475)(15808,3.48717)(15912,3.53231)(16016,3.56931)(16120,3.61502)(16224,3.65114)(16328,3.694)(16432,3.74637)(16536,3.79424)(16640,3.83676)(16744,3.88711)(16848,3.93319)(16952,3.97898)(17056,4.02701)(17160,4.07119)(17264,4.12715)(17368,4.17232)(17472,4.22443)(17576,4.27788)(17680,4.33607)(17784,4.38027)(17888,4.43319)(17992,4.48196)(18096,4.5326)(18200,4.58809)(18304,4.65024)(18408,4.70081)(18512,4.75053)(18616,4.82103)(18720,4.8681)(18824,4.92623)(18928,4.99024)(19032,5.05068)(19136,5.12328)(19240,5.18995)(19344,5.26404)(19448,5.3297)(19552,5.40014)(19656,5.47749)(19760,5.54298)(19864,5.61174)(19968,5.70231)(20072,5.78551)(20176,5.88395)(20280,6.0473)(20384,6.3145)(20488,6.70579)(20592,7.30064)
	};
	\addplot [mark=x,only marks] coordinates {
		(8820,1.52854)(9947,1.76019)(13695,2.73423)(14504,3.01951)
	};
	\node [anchor=south] at (axis cs: 8820,1.52854) {picture};
	\node [anchor=south] at (axis cs: 9947,1.76019) {story};
	\node [anchor=south] at (axis cs: 13695,2.73423) {paint};
	\node [anchor=south] at (axis cs: 14504,3.01951) {words};
	\end{axis}
\end{tikzpicture}
\caption{\textsc{dimension: tone}}
\end{subfigure}
\centering
\begin{subfigure} [t] {0.3 \textwidth}
\begin{tikzpicture} [scale=0.5]
	\begin{axis} [xlabel=rank,ylabel=PMI,y label style={color=white}]
	\addplot [mark=none] coordinates {
		(0,0.03484)(58,0.09256)(116,0.12683)(174,0.1543)(232,0.17249)(290,0.19221)(348,0.21423)(406,0.23361)(464,0.25217)(522,0.2728)(580,0.28898)(638,0.30378)(696,0.32327)(754,0.33943)(812,0.35621)(870,0.37557)(928,0.3898)(986,0.40192)(1044,0.41563)(1102,0.43014)(1160,0.44149)(1218,0.45305)(1276,0.46988)(1334,0.48637)(1392,0.50335)(1450,0.51913)(1508,0.53302)(1566,0.54636)(1624,0.56162)(1682,0.57734)(1740,0.59143)(1798,0.60665)(1856,0.61916)(1914,0.6371)(1972,0.65116)(2030,0.66306)(2088,0.67659)(2146,0.69151)(2204,0.70535)(2262,0.71855)(2320,0.73193)(2378,0.74266)(2436,0.75729)(2494,0.77546)(2552,0.78633)(2610,0.80012)(2668,0.8169)(2726,0.83537)(2784,0.85355)(2842,0.86732)(2900,0.88322)(2958,0.89446)(3016,0.90613)(3074,0.9231)(3132,0.94014)(3190,0.95222)(3248,0.96711)(3306,0.9849)(3364,1.00205)(3422,1.01845)(3480,1.03507)(3538,1.04852)(3596,1.06274)(3654,1.07879)(3712,1.09558)(3770,1.11291)(3828,1.12954)(3886,1.14094)(3944,1.1583)(4002,1.17565)(4060,1.19193)(4118,1.20659)(4176,1.22144)(4234,1.24045)(4292,1.25819)(4350,1.27497)(4408,1.29346)(4466,1.30894)(4524,1.32576)(4582,1.34537)(4640,1.36236)(4698,1.37802)(4756,1.40092)(4814,1.41812)(4872,1.437)(4930,1.45362)(4988,1.47054)(5046,1.48794)(5104,1.50497)(5162,1.5235)(5220,1.54608)(5278,1.56993)(5336,1.58838)(5394,1.60762)(5452,1.6294)(5510,1.64545)(5568,1.66758)(5626,1.68628)(5684,1.70118)(5742,1.71768)(5800,1.73827)(5858,1.75865)(5916,1.7775)(5974,1.7962)(6032,1.81993)(6090,1.8409)(6148,1.85916)(6206,1.88354)(6264,1.90946)(6322,1.92976)(6380,1.95376)(6438,1.97916)(6496,1.99664)(6554,2.01574)(6612,2.04199)(6670,2.0636)(6728,2.08629)(6786,2.10599)(6844,2.13444)(6902,2.16308)(6960,2.18992)(7018,2.21645)(7076,2.24696)(7134,2.27151)(7192,2.30551)(7250,2.33204)(7308,2.3618)(7366,2.39347)(7424,2.42329)(7482,2.45121)(7540,2.48118)(7598,2.51688)(7656,2.54933)(7714,2.57875)(7772,2.60519)(7830,2.64168)(7888,2.67553)(7946,2.70494)(8004,2.73943)(8062,2.77279)(8120,2.81042)(8178,2.84329)(8236,2.87581)(8294,2.91297)(8352,2.95732)(8410,3.00346)(8468,3.03243)(8526,3.06709)(8584,3.09843)(8642,3.1325)(8700,3.16315)(8758,3.20404)(8816,3.24392)(8874,3.28262)(8932,3.31794)(8990,3.3667)(9048,3.41114)(9106,3.46778)(9164,3.51691)(9222,3.56967)(9280,3.6222)(9338,3.67966)(9396,3.71954)(9454,3.75608)(9512,3.80428)(9570,3.86122)(9628,3.91146)(9686,3.97889)(9744,4.04215)(9802,4.08914)(9860,4.1583)(9918,4.22125)(9976,4.28521)(10034,4.36149)(10092,4.43298)(10150,4.51891)(10208,4.59721)(10266,4.68116)(10324,4.76401)(10382,4.86804)(10440,4.94818)(10498,5.02824)(10556,5.12139)(10614,5.23003)(10672,5.33196)(10730,5.42782)(10788,5.49519)(10846,5.62498)(10904,5.7597)(10962,5.91086)(11020,6.03144)(11078,6.13471)(11136,6.25668)(11194,6.47501)(11252,6.62009)(11310,6.79369)(11368,6.97399)(11426,7.55902)
	};
	\addplot [mark=x,only marks] coordinates {
		(2845,0.86784)(4192,1.22602)(7830,2.64168)(8654,3.13637)
	};
	\node [anchor=south] at (axis cs: 2845,0.86784) {paint};
	\node [anchor=south] at (axis cs: 4192,1.22602) {picture};
	\node [anchor=south] at (axis cs: 7830,2.64168) {words};
	\node [anchor=south] at (axis cs: 8654,3.13637) {story};
	\end{axis}
\end{tikzpicture}
\caption{\textsc{dimension: relate}}
\end{subfigure}
\centering
\begin{subfigure} [t] {0.3 \textwidth}
\begin{tikzpicture} [scale=0.5]
	\begin{axis} [xlabel=rank,ylabel=PMI,y label style={color=white}]
	\addplot [mark=none] coordinates {
		(0,0.03273)(125,0.1606)(250,0.20302)(375,0.23629)(500,0.2665)(625,0.28849)(750,0.30879)(875,0.32835)(1000,0.34637)(1125,0.36376)(1250,0.3802)(1375,0.39824)(1500,0.41362)(1625,0.42889)(1750,0.44542)(1875,0.46026)(2000,0.47525)(2125,0.48734)(2250,0.49937)(2375,0.51631)(2500,0.53083)(2625,0.54165)(2750,0.55431)(2875,0.56641)(3000,0.57824)(3125,0.59035)(3250,0.60052)(3375,0.6126)(3500,0.62472)(3625,0.63737)(3750,0.64756)(3875,0.66147)(4000,0.67301)(4125,0.68296)(4250,0.69439)(4375,0.70491)(4500,0.71578)(4625,0.72684)(4750,0.74052)(4875,0.75045)(5000,0.76208)(5125,0.7743)(5250,0.7867)(5375,0.79734)(5500,0.80914)(5625,0.82082)(5750,0.83392)(5875,0.84701)(6000,0.85971)(6125,0.8698)(6250,0.88137)(6375,0.89324)(6500,0.90498)(6625,0.91746)(6750,0.92985)(6875,0.94286)(7000,0.957)(7125,0.96919)(7250,0.98318)(7375,0.99664)(7500,1.01023)(7625,1.02503)(7750,1.03727)(7875,1.0538)(8000,1.06708)(8125,1.07977)(8250,1.09329)(8375,1.11047)(8500,1.12395)(8625,1.13854)(8750,1.15422)(8875,1.1667)(9000,1.18027)(9125,1.19516)(9250,1.20924)(9375,1.22456)(9500,1.2406)(9625,1.25802)(9750,1.27524)(9875,1.29084)(10000,1.3043)(10125,1.32238)(10250,1.34042)(10375,1.35878)(10500,1.37676)(10625,1.39404)(10750,1.40928)(10875,1.42932)(11000,1.44773)(11125,1.46637)(11250,1.48536)(11375,1.50393)(11500,1.52583)(11625,1.545)(11750,1.56686)(11875,1.58171)(12000,1.59936)(12125,1.61757)(12250,1.64016)(12375,1.66148)(12500,1.679)(12625,1.70142)(12750,1.72236)(12875,1.74163)(13000,1.76267)(13125,1.78224)(13250,1.80531)(13375,1.82651)(13500,1.84986)(13625,1.87226)(13750,1.89482)(13875,1.91662)(14000,1.93827)(14125,1.96179)(14250,1.98619)(14375,2.00954)(14500,2.03377)(14625,2.05946)(14750,2.08612)(14875,2.10769)(15000,2.13051)(15125,2.15538)(15250,2.17846)(15375,2.20398)(15500,2.22996)(15625,2.25651)(15750,2.28459)(15875,2.31469)(16000,2.3448)(16125,2.37496)(16250,2.40686)(16375,2.43231)(16500,2.45893)(16625,2.48849)(16750,2.52137)(16875,2.54756)(17000,2.57838)(17125,2.60944)(17250,2.63721)(17375,2.6702)(17500,2.70731)(17625,2.74256)(17750,2.77815)(17875,2.81263)(18000,2.85037)(18125,2.88739)(18250,2.92207)(18375,2.96425)(18500,2.99885)(18625,3.03911)(18750,3.07616)(18875,3.11304)(19000,3.16001)(19125,3.19865)(19250,3.24116)(19375,3.28677)(19500,3.32941)(19625,3.38385)(19750,3.43001)(19875,3.47647)(20000,3.52317)(20125,3.57427)(20250,3.61102)(20375,3.65938)(20500,3.70659)(20625,3.75926)(20750,3.81323)(20875,3.86963)(21000,3.9181)(21125,3.97486)(21250,4.03367)(21375,4.09342)(21500,4.14608)(21625,4.20794)(21750,4.26051)(21875,4.31379)(22000,4.37091)(22125,4.43312)(22250,4.48889)(22375,4.55074)(22500,4.61372)(22625,4.68382)(22750,4.74623)(22875,4.80731)(23000,4.87821)(23125,4.94817)(23250,5.01427)(23375,5.08543)(23500,5.16265)(23625,5.22535)(23750,5.30602)(23875,5.38501)(24000,5.45729)(24125,5.55995)(24250,5.65854)(24375,5.80042)(24500,6.06459)(24625,6.38988)(24750,6.8489)(24875,8.11699)
	};
	\addplot [mark=x,only marks] coordinates {
		(2769,0.5561)(6690,0.92441)(20895,3.87616)(21475,4.13299)
	};
	\node [anchor=south] at (axis cs: 2769,0.5561) {words};
	\node [anchor=south] at (axis cs: 6690,0.92441) {story};
	\node [anchor=south] at (axis cs: 20895,3.87616) {paint};
	\node [anchor=south] at (axis cs: 21475,4.13299) {picture};
	\end{axis}
\end{tikzpicture}
\caption{\textsc{dimension: gray}}
\end{subfigure}
\caption{Selected Analogical Dimensions.  These are 3 of 20 dimensions that geometrically map the analogy \emph{picture} is to \emph{paint} as \emph{story} is to \emph{words}, presented as histograms ranking word-vectors with non-zero PMI values and highlighting the positions of the analogical components along this curve.  Of particular note is the way that analogically oriented word pairs tend to cluster.}
\label{FIG:histograms}
\end{figure}
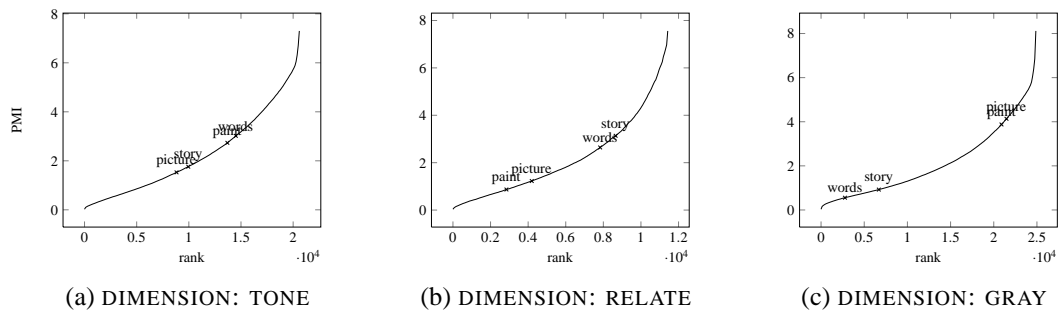

To further explore the nature of the subspaces where these conceptual relationships are mapped, the previously discussed analogy \emph{picture} is to \emph{paint} as \emph{story} is to \emph{words} has been examined, and a selection of some of the dimensions ultimately used to map the relationship are depicted in Figure~\ref{FIG:histograms}.  The thing to note about the way that the analogy plays out along these dimensions is the way that the relative PMI values of the word-vectors in question cluster into pairs.  These pairs will consist of iterations of a given word-vector being coupled with one its two analogues, and over the course of several dimensions a given pair will be observed relatively towards both the high end or the low end of the spectrum of values.

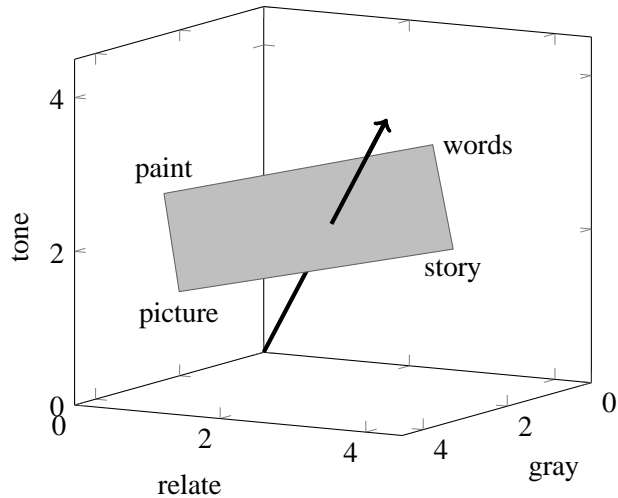
\begin{figure}
\centering
\begin{tikzpicture}
\begin{axis}[view={120}{10},xmin=0,xmax=4.5,ymin=0,ymax=4.5,zmin=0,zmax=4.5,xlabel=gray,ylabel=relate,zlabel=tone,colormap/blackwhite]
\addplot3 [color=black,->,ultra thick] coordinates {(0,0,0) (4,4,4)};
\addplot3[patch,patch type=rectangle,color=lightgray,fill opacity=1] coordinates {
 	(3.876,0.868,2.734) (4.133,1.226,1.528) (0.924,3.136,1.760) (0.556,2.642,3.019)
};
\node [anchor=south] at (axis cs: 3.876,0.868,2.734) {paint};
\node [anchor=north] at (axis cs: 4.133,1.226,1.528) {picture};
\node [anchor=north] at (axis cs: 0.924,3.136,1.760) {story};
\node [anchor=west] at (axis cs: 0.556,2.642,3.019) {words};
\addplot3 [color=black,->,ultra thick] coordinates {(2.2,2.2,2.2) (4,4,4)};
\end{axis}
\end{tikzpicture}
\caption{An Analogy in Space. When the analogue values of the three dimensions described in Figure~\ref{FIG:histograms} are plotted, the analogy itself emerges as a parallelogram situated obliquely in the centre of the positive region of the space.}
\label{FIG:subspace}
\end{figure}

The net geometric result of this tendency is that the parallelogram delimiting the analogy is pushed into an orientation where it sits more or less obliquely near the centre of the positive region of the subspace, approximately perpendicular to a centre line extending from the origin, as illustrated in Figure~\ref{FIG:subspace}.  The distinguishing features of this characteristic analogical geometry -- its flatness, its centrality, and its obliqueness -- are hardly surprising when considered in the context of the properties of the dimensions that give the edges of the analogy their fundamental trait of parallelism.  Nonetheless, the insight gained from pursuing the geometry of these intuitions might be very valuable tools in the geometric description of analogy.  This insight might be pushed in at least two different directions.  For one thing, given three of the four legs of an analogy, we now have a hunch about the way those three word-vectors should sit in a subspace that will complete the analogy in a satisfactory way.  For another thing, given a particular context delineated in terms of co-occurrence dimensions, we now have a mechanism for beginning to look for regions of the space that can bear an analogical conceptual relationship.

\section{Final Thoughts}
What has been presented here is in effect the beginning of a work-in-progress.  The exploration of the way that a dynamic distributional semantic space might be used to project analogically productive subspaces in an online, context sensitive way has not yet yielded a methodology for actually performing, for instance, an analogy completion task.  What has been demonstrated, however, is the way that a robustly geometrical approach to corpus linguistics can bear some strong results.  Rather than having a space simply for the sake of situating words in relation to one another, we have proposed building spaces which are themselves full of meaning and the potential for interpretation.  By situating words in a dynamic space, the problems inherent in the construction of highly informative systems of interactive symbols becomes the natural outcome of spaces which are necessarily replete with interrelationships.  So, finally, rather than needing to try to model conceptualisation as a catastrophically unwieldy network of encyclopedic relationships, the cognitive situation inherent in conceptualisation becomes, like the world itself, thoroughly geometric.

\bibliographystyle{eptcs}
\bibliography{Cite}

\begin{thebibliography}{10}
\providecommand{\bibitemdeclare}[2]{}
\providecommand{\surnamestart}{}
\providecommand{\surnameend}{}
\providecommand{\urlprefix}{Available at }
\providecommand{\url}[1]{\texttt{#1}}
\providecommand{\href}[2]{\texttt{#2}}
\providecommand{\urlalt}[2]{\href{#1}{#2}}
\providecommand{\doi}[1]{doi:\urlalt{http://dx.doi.org/#1}{#1}}
\providecommand{\bibinfo}[2]{#2}

\bibitemdeclare{inproceedings}{AgresEA2015}
\bibitem{AgresEA2015}
\bibinfo{author}{Kat \surnamestart Agres\surnameend}, \bibinfo{author}{Stephen
  \surnamestart McGregor\surnameend}, \bibinfo{author}{Matthew \surnamestart
  Purver\surnameend} \& \bibinfo{author}{Geraint \surnamestart
  Wiggins\surnameend} (\bibinfo{year}{2015}):
  \emph{\bibinfo{title}{Conceptualising Creativity: From Distributional
  Semantics to Conceptual Spaces}}.
\newblock In: {\sl \bibinfo{booktitle}{Proceedings of the 6th International
  Conference on Computational Creativity}}, \bibinfo{address}{Park City, UT}.
\newblock
  \urlprefix\url{http://computationalcreativity.net/iccc2015/proceedings/5\_4Agres.pdf}.

\bibitemdeclare{book}{Aristotle1895}
\bibitem{Aristotle1895}
\bibinfo{author}{\surnamestart Aristotle\surnameend} (\bibinfo{year}{1895}):
  \emph{\bibinfo{title}{The Poetics}}.
\newblock \bibinfo{publisher}{Macmillan and Co}, \bibinfo{address}{London}.

\bibitemdeclare{inproceedings}{BaroniEA2010}
\bibitem{BaroniEA2010}
\bibinfo{author}{Marco \surnamestart Baroni\surnameend} \&
  \bibinfo{author}{Roberto \surnamestart Zamparelli\surnameend}
  (\bibinfo{year}{2010}): \emph{\bibinfo{title}{Nouns are vectors, adjectives
  are matrices: Representing adjective-noun constructions in semantic space}}.
\newblock In: {\sl \bibinfo{booktitle}{Proceedings of the 2010 Conference on
  Empirical Methods in Natural Language Processing}}, pp.
  \bibinfo{pages}{1183--1193}.
\newblock \urlprefix\url{http://dl.acm.org/citation.cfm?id=1870658.1870773}.

\bibitemdeclare{inbook}{Barsalou1993}
\bibitem{Barsalou1993}
\bibinfo{author}{Lawrence~W. \surnamestart Barsalou\surnameend}
  (\bibinfo{year}{1993}): \emph{\bibinfo{title}{Theories of Memory}}, chapter
  \bibinfo{chapter}{Flexibility, Structure, and Linguistic Vagary in Concepts:
  Manifestations of a Compositional System of Perceptual Symbols}.
\newblock \bibinfo{publisher}{Lawrence Erlbaum Associates},
  \bibinfo{address}{Hove}.

\bibitemdeclare{article}{BengioEA2003}
\bibitem{BengioEA2003}
\bibinfo{author}{Yoshua \surnamestart Bengio\surnameend},
  \bibinfo{author}{Réjean \surnamestart Ducharme\surnameend},
  \bibinfo{author}{Pascal \surnamestart Vincent\surnameend} \&
  \bibinfo{author}{Christian \surnamestart Jauvin\surnameend}
  (\bibinfo{year}{2003}): \emph{\bibinfo{title}{A Neural Probabilistic Language
  Model}}.
\newblock {\sl \bibinfo{journal}{Journal of Machine Learning Research}}
  \bibinfo{volume}{3}, pp. \bibinfo{pages}{1137--1155}.
\newblock \urlprefix\url{http://dl.acm.org/citation.cfm?id=944919.944966}.

\bibitemdeclare{incollection}{BordesEA2013}
\bibitem{BordesEA2013}
\bibinfo{author}{Antoine \surnamestart Bordes\surnameend},
  \bibinfo{author}{Nicolas \surnamestart Usunier\surnameend},
  \bibinfo{author}{Alberto \surnamestart Garcia-Duran\surnameend},
  \bibinfo{author}{Jason \surnamestart Weston\surnameend} \&
  \bibinfo{author}{Oksana \surnamestart Yakhnenko\surnameend}
  (\bibinfo{year}{2013}): \emph{\bibinfo{title}{Translating Embeddings for
  Modeling Multi-relational Data}}.
\newblock In \bibinfo{editor}{C.~J.~C. \surnamestart Burges\surnameend},
  \bibinfo{editor}{L.~\surnamestart Bottou\surnameend},
  \bibinfo{editor}{M.~\surnamestart Welling\surnameend},
  \bibinfo{editor}{Z.~\surnamestart Ghahramani\surnameend} \&
  \bibinfo{editor}{K.~Q. \surnamestart Weinberger\surnameend}, editors: {\sl
  \bibinfo{booktitle}{Advances in Neural Information Processing Systems 26}},
  \bibinfo{publisher}{Curran Associates, Inc.}, pp.
  \bibinfo{pages}{2787--2795}.

\bibitemdeclare{incollection}{Clark2015}
\bibitem{Clark2015}
\bibinfo{author}{Stephen \surnamestart Clark\surnameend}
  (\bibinfo{year}{2015}): \emph{\bibinfo{title}{Vector Space Models of Lexical
  Meaning}}.
\newblock In \bibinfo{editor}{Shalom \surnamestart Lappin\surnameend} \&
  \bibinfo{editor}{Chris \surnamestart Fox\surnameend}, editors: {\sl
  \bibinfo{booktitle}{The Handbook of Contemporary Semantic Theory}},
  \bibinfo{publisher}{Wiley-Blackwell}, \doi{10.1002/9781118882139.ch16}.

\bibitemdeclare{inproceedings}{CollobertEA2008}
\bibitem{CollobertEA2008}
\bibinfo{author}{Ronan \surnamestart Collobert\surnameend} \&
  \bibinfo{author}{Jason \surnamestart Weston\surnameend}
  (\bibinfo{year}{2008}): \emph{\bibinfo{title}{A Unified Architecture for
  Natural Language Processing: Deep Neural Networks with Multitask Learning}}.
\newblock In: {\sl \bibinfo{booktitle}{Proceedings of the 25 th International
  Conference on Machine Learning}}, \doi{10.1145/1390156.1390177}.

\bibitemdeclare{incollection}{Cooper2012}
\bibitem{Cooper2012}
\bibinfo{author}{Robin \surnamestart Cooper\surnameend} (\bibinfo{year}{2012}):
  \emph{\bibinfo{title}{Type Theory and Semantics in Flux}}.
\newblock In \bibinfo{editor}{Ruth \surnamestart Kempson\surnameend},
  \bibinfo{editor}{Tim \surnamestart Fernando\surnameend} \&
  \bibinfo{editor}{Nicholas \surnamestart Asher\surnameend}, editors: {\sl
  \bibinfo{booktitle}{Philosophy of Linguistics}},
  \bibinfo{publisher}{Elsevier}, \doi{10.1016/B978-0-444-51747-0.50009-3}.

\bibitemdeclare{incollection}{Davidson1969}
\bibitem{Davidson1969}
\bibinfo{author}{Donald \surnamestart Davidson\surnameend}
  (\bibinfo{year}{1969}): \emph{\bibinfo{title}{The Individuation of Events}}.
\newblock In \bibinfo{editor}{Nicholas \surnamestart Rescher\surnameend},
  editor: {\sl \bibinfo{booktitle}{Essays in Honor of Carl G. Hempel}}, pp.
  \bibinfo{pages}{216--234}, \doi{10.1007/978-94-017-1466-2\_11}.

\bibitemdeclare{incollection}{Dennett1984}
\bibitem{Dennett1984}
\bibinfo{author}{Daniel~C. \surnamestart Dennett\surnameend}
  (\bibinfo{year}{1984}): \emph{\bibinfo{title}{Cognitive Wheels: The Frame
  Problem of AI}}.
\newblock In \bibinfo{editor}{Christopher \surnamestart Hookway\surnameend},
  editor: {\sl \bibinfo{booktitle}{Minds, Machines, and Evolution:
  Philosophical Studies}}, \bibinfo{publisher}{Cambridge University Press}, pp.
  \bibinfo{pages}{129--151}.

\bibitemdeclare{inproceedings}{DerracEA2014}
\bibitem{DerracEA2014}
\bibinfo{author}{Joaqu{\'i}n \surnamestart Derrac\surnameend} \&
  \bibinfo{author}{Steven \surnamestart Schockaert\surnameend}
  (\bibinfo{year}{2014}): \emph{\bibinfo{title}{Characterising Semantic
  Relatedness Using Interpretable Directions in Conceptual Spaces}}.
\newblock In: {\sl \bibinfo{booktitle}{2nd European Conference on Artificial
  Intelligence}}, pp. \bibinfo{pages}{243--248},
  \doi{10.3233/978-1-61499-419-0-243}.

\bibitemdeclare{article}{FodorEA1988}
\bibitem{FodorEA1988}
\bibinfo{author}{Jerry~A. \surnamestart Fodor\surnameend} \&
  \bibinfo{author}{Zenon~W. \surnamestart Pylyshyn\surnameend}
  (\bibinfo{year}{1988}): \emph{\bibinfo{title}{Connectionism and Cognitive
  Architecture: A Critical Analysis}}.
\newblock {\sl \bibinfo{journal}{Cognition}}
  \bibinfo{volume}{28}(\bibinfo{number}{1-2}), pp. \bibinfo{pages}{3--71},
  \doi{10.1016/0010-0277(88)90031-5}.

\bibitemdeclare{book}{Gardenfors2000}
\bibitem{Gardenfors2000}
\bibinfo{author}{Peter \surnamestart G{\"a}rdenfors\surnameend}
  (\bibinfo{year}{2000}): \emph{\bibinfo{title}{Conceptual Space: The Geometry
  of Thought}}.
\newblock \bibinfo{publisher}{The MIT Press}, \bibinfo{address}{Cambridge, MA}.

\bibitemdeclare{book}{Gardenfors2014}
\bibitem{Gardenfors2014}
\bibinfo{author}{Peter \surnamestart G{\"a}rdenfors\surnameend}
  (\bibinfo{year}{2014}): \emph{\bibinfo{title}{The Geometry of Meaning:
  Semantics Based on Conceptual Spaces}}.
\newblock \bibinfo{publisher}{The MIT Press}, \bibinfo{address}{Cambridge, MA}.

\bibitemdeclare{incollection}{Haspelmath2003}
\bibitem{Haspelmath2003}
\bibinfo{author}{Martin \surnamestart Haspelmath\surnameend}
  (\bibinfo{year}{2003}): \emph{\bibinfo{title}{The geometry of grammatical
  meaning: semantic maps and cross-linguistic comparison}}.
\newblock In \bibinfo{editor}{Michael \surnamestart Tomasello\surnameend},
  editor: {\sl \bibinfo{booktitle}{The new psychology of language: cognitive
  and functional approaches to language structure, vol. 2}},
  \bibinfo{publisher}{Lawrence Erlbaum}, \bibinfo{address}{Mahwah, New
  Jersey/Londo}, pp. \bibinfo{pages}{211--242}.
\newblock \urlprefix\url{http://wwwstaff.eva.mpg.de/~haspelmt/SemMaps.pdf}.

\bibitemdeclare{book}{Hesse1963}
\bibitem{Hesse1963}
\bibinfo{author}{Mary~B. \surnamestart Hesse\surnameend}
  (\bibinfo{year}{1963}): \emph{\bibinfo{title}{Models and Analogies in
  Science}}.
\newblock \bibinfo{publisher}{Sheed and Ward}, \bibinfo{address}{New York}.

\bibitemdeclare{inproceedings}{LevyEA2014}
\bibitem{LevyEA2014}
\bibinfo{author}{Omer \surnamestart Levy\surnameend} \& \bibinfo{author}{Yoav
  \surnamestart Goldberg\surnameend} (\bibinfo{year}{2014}):
  \emph{\bibinfo{title}{Linguistic Regularities in Sparse and Explicit Word
  Representations}}.
\newblock In: {\sl \bibinfo{booktitle}{Eighteenth Conference on Computational
  Natural Language Learning}}, \doi{10.3115/v1/W14-1618}.

\bibitemdeclare{article}{McGregorEA2015c}
\bibitem{McGregorEA2015c}
\bibinfo{author}{Stephen \surnamestart McGregor\surnameend},
  \bibinfo{author}{Kat \surnamestart Agres\surnameend},
  \bibinfo{author}{Matthew \surnamestart Purver\surnameend} \&
  \bibinfo{author}{Geraint \surnamestart Wiggins\surnameend}
  (\bibinfo{year}{2015}): \emph{\bibinfo{title}{From Distributional Semantics
  to Conceptual Spaces: A Novel Computational Method for Concept Creation}}.
\newblock {\sl \bibinfo{journal}{Journal of Artificial General Intelligence}},
  \doi{10.1515/jagi-2015-0004}.

\bibitemdeclare{inproceedings}{MikolovEA2013b}
\bibitem{MikolovEA2013b}
\bibinfo{author}{Tomas \surnamestart Mikolov\surnameend}, \bibinfo{author}{Kai
  \surnamestart Chen\surnameend}, \bibinfo{author}{Greg \surnamestart
  Corrado\surnameend} \& \bibinfo{author}{Jeffrey \surnamestart
  Dean\surnameend} (\bibinfo{year}{2013}): \emph{\bibinfo{title}{Efficient
  Estimation of Word Representations in Vector Space}}.
\newblock In: {\sl \bibinfo{booktitle}{Proceedings of ICLR Workshop}}.
\newblock \urlprefix\url{http://arxiv.org/abs/1301.3781}.

\bibitemdeclare{inproceedings}{MikolovEA2013}
\bibitem{MikolovEA2013}
\bibinfo{author}{Tomas \surnamestart Mikolov\surnameend}, \bibinfo{author}{Wen
  \surnamestart tau Yih\surnameend} \& \bibinfo{author}{Geoffrey \surnamestart
  Zweig\surnameend} (\bibinfo{year}{2013}): \emph{\bibinfo{title}{Linguistic
  Regularities in Continuous Space Word Representations}}.
\newblock In: {\sl \bibinfo{booktitle}{Proceedings of the 2013 Conference of
  the North American Chapter of the Association for Computational Linguistics:
  Human Language Technologies}}, pp. \bibinfo{pages}{246--251}.
\newblock \urlprefix\url{http://www.aclweb.org/anthology/N13-1090}.

\bibitemdeclare{incollection}{Montague1974}
\bibitem{Montague1974}
\bibinfo{author}{Richard \surnamestart Montague\surnameend}
  (\bibinfo{year}{1974}): \emph{\bibinfo{title}{English as a Formal Language}}.
\newblock In \bibinfo{editor}{Richard~H. \surnamestart Thompson\surnameend},
  editor: {\sl \bibinfo{booktitle}{Formal Philosophy: selected papers of
  Richard Montague}}, \bibinfo{publisher}{Yale University Press},
  \bibinfo{address}{New Haven, CT}.

\bibitemdeclare{inproceedings}{PenningtonEA2014}
\bibitem{PenningtonEA2014}
\bibinfo{author}{Jeffrey \surnamestart Pennington\surnameend},
  \bibinfo{author}{Richard \surnamestart Socher\surnameend} \&
  \bibinfo{author}{Christopher~D. \surnamestart Manning\surnameend}
  (\bibinfo{year}{2014}): \emph{\bibinfo{title}{GloVe: Global Vectors for Word
  Representation}}.
\newblock In: {\sl \bibinfo{booktitle}{Conference on Empirical Methods in
  Natural Language Processing}}, \doi{10.3115/v1/D14-1162}.

\bibitemdeclare{inproceedings}{Rimell2014}
\bibitem{Rimell2014}
\bibinfo{author}{Laura \surnamestart Rimell\surnameend} (\bibinfo{year}{2014}):
  \emph{\bibinfo{title}{Distributional Lexical Entailment by Topic Coherence}}.
\newblock In: {\sl \bibinfo{booktitle}{Proceedings of the 14th Conference of
  the European Chapter of the Association for Computational Linguistics}},
  \bibinfo{address}{Gothenburg}, \doi{10.3115/v1/E14-1054}.

\bibitemdeclare{inproceedings}{SaltonEA1975}
\bibitem{SaltonEA1975}
\bibinfo{author}{G.~\surnamestart Salton\surnameend},
  \bibinfo{author}{A.~\surnamestart Wong\surnameend} \& \bibinfo{author}{C.~S.
  \surnamestart Yang\surnameend} (\bibinfo{year}{1975}):
  \emph{\bibinfo{title}{A Vector Space Model for Automatic Indexing}}.
\newblock In: {\sl \bibinfo{booktitle}{Proceedings of the 12th ACM SIGIR
  Conference}}, pp. \bibinfo{pages}{137--150}, \doi{10.1145/361219.361220}.

\bibitemdeclare{inproceedings}{Schutze1992}
\bibitem{Schutze1992}
\bibinfo{author}{Hinrich \surnamestart Sch\"{u}tze\surnameend}
  (\bibinfo{year}{1992}): \emph{\bibinfo{title}{Dimensions of Meaning}}.
\newblock In: {\sl \bibinfo{booktitle}{Proceedings of the 1992 ACM/IEEE
  conference on Supercomputing}}, pp. \bibinfo{pages}{787--796},
  \doi{10.1109/SUPERC.1992.236684}.

\bibitemdeclare{book}{SperberEA1995}
\bibitem{SperberEA1995}
\bibinfo{author}{Dan \surnamestart Sperber\surnameend} \&
  \bibinfo{author}{Deirdre \surnamestart Wilson\surnameend}
  (\bibinfo{year}{1995}): \emph{\bibinfo{title}{Relevance: Communication and
  Cognition}}, \bibinfo{edition}{2nd} edition.
\newblock \bibinfo{publisher}{Blackwell}.

\bibitemdeclare{article}{TurneyEA2010}
\bibitem{TurneyEA2010}
\bibinfo{author}{Peter~D. \surnamestart Turney\surnameend} \&
  \bibinfo{author}{Patrick \surnamestart Patel\surnameend}
  (\bibinfo{year}{2010}): \emph{\bibinfo{title}{From Frequency to Meaning:
  Vector Space Models of Semantics}}.
\newblock {\sl \bibinfo{journal}{Journal of Artificial Intelligence Research}}
  \bibinfo{volume}{37}, pp. \bibinfo{pages}{141--188}, \doi{10.1613/jair.2934}.

\bibitemdeclare{book}{Widdows2004}
\bibitem{Widdows2004}
\bibinfo{author}{Dominic \surnamestart Widdows\surnameend}
  (\bibinfo{year}{2004}): \emph{\bibinfo{title}{Geometry and Meaning}}.
\newblock \bibinfo{publisher}{CSLI Publications}, \bibinfo{address}{Stanford,
  CA}.

\end{thebibliography}
\end{document}